\documentclass{ws-procs11x85}
\usepackage{ws-procs-thm}           
\usepackage[most]{tcolorbox}
\usepackage{xcolor}

\begin{document}

\title{TrueImage: A Machine Learning Algorithm to Improve the Quality of Telehealth Photos}

\author{Kailas Vodrahalli$^{1,\dag,\ast}$, Roxana Daneshjou$^{2,3,\dag,\ast}$, Roberto A Novoa$^{2,4}$, Albert Chiou$^{2}$, Justin M Ko$^{2}$, and James Zou$^{1,3,\dag}$}

\address{$^{1}$Department of Electrical Engineering, Stanford University,\\
Stanford, CA 94305, USA\\
$^{2}$Department of Dermatology, Stanford University School of Medicine,\\
Redwood City, CA 94063, USA\\
$^{3}$Department of Biomedical Data Science, Stanford University School of Medicine,\\
Stanford, CA 94305, USA\\
$^{4}$Department of Pathology, Stanford University School of Medicine,\\
Stanford, CA 94305, USA\\
$^{\dag}$ Correspondence can be addressed to: kailasv@stanford.edu, roxanad@stanford.edu, jamesz@stanford.edu \\
$^{\ast}$ These authors contributed equally
}

\begin{abstract}
Telehealth is an increasingly critical component of the health care ecosystem, especially due to the COVID-19 pandemic. Rapid adoption of telehealth has exposed limitations in the existing infrastructure. In this paper, we study and highlight photo quality as a major challenge in the telehealth workflow. We focus on teledermatology, where photo quality is particularly important; the framework proposed here can be generalized to other health domains.    
For telemedicine, dermatologists request that patients submit images of their lesions for assessment. However, these images are often of insufficient quality to make a clinical diagnosis since patients do not have experience taking clinical photos.  A clinician has to manually triage poor quality images and request new images to be submitted, leading to wasted time for both the clinician and the patient. We propose an automated image assessment machine learning pipeline, TrueImage, to detect poor quality dermatology photos and to guide patients in taking better photos. Our experiments indicate that TrueImage can reject $\sim$$50\%$ of the sub-par quality images, while retaining $\sim$$80\%$ of good quality images patients send in, despite heterogeneity and limitations in the training data. These promising results suggest that our solution is feasible and can improve the quality of teledermatology care.
\end{abstract}


\copyrightinfo{\copyright\ 2020 The Authors. Open Access chapter published by World Scientific Publishing Company and distributed under the terms of the Creative Commons Attribution Non-Commercial (CC BY-NC) 4.0 License.}

\section{Introduction}

Due to the SARS-CoV-2 (COVID-19) pandemic, many hospitals have rapidly transitioned patient visits to video conference calls on a digital platform to limit exposure for both patients and healthcare workers. Although these digital visits have some limitations, they have recently accounted for more than $10\%$ of all visits in the US, corresponding to more than an $10000\%$ increase since February 2020 \cite{mehrotra2020impact}.

The rapid adoption of telehealth has unearthed substantial challenges. For example, productive teledermatology visits require high clinical quality images of the area of concern; however, video call platforms do not have sufficient imaging resolution for diagnosis.
In teledermatology, a clinician will often request patients to send in photos of their lesions or rash ahead of time. The clinician will use these images for assessing the patient's condition and use the digital platform of the visit to communicate with a patient rather than for making assessments.

\begin{figure}[h]
\centerline{\includegraphics[width=12cm]{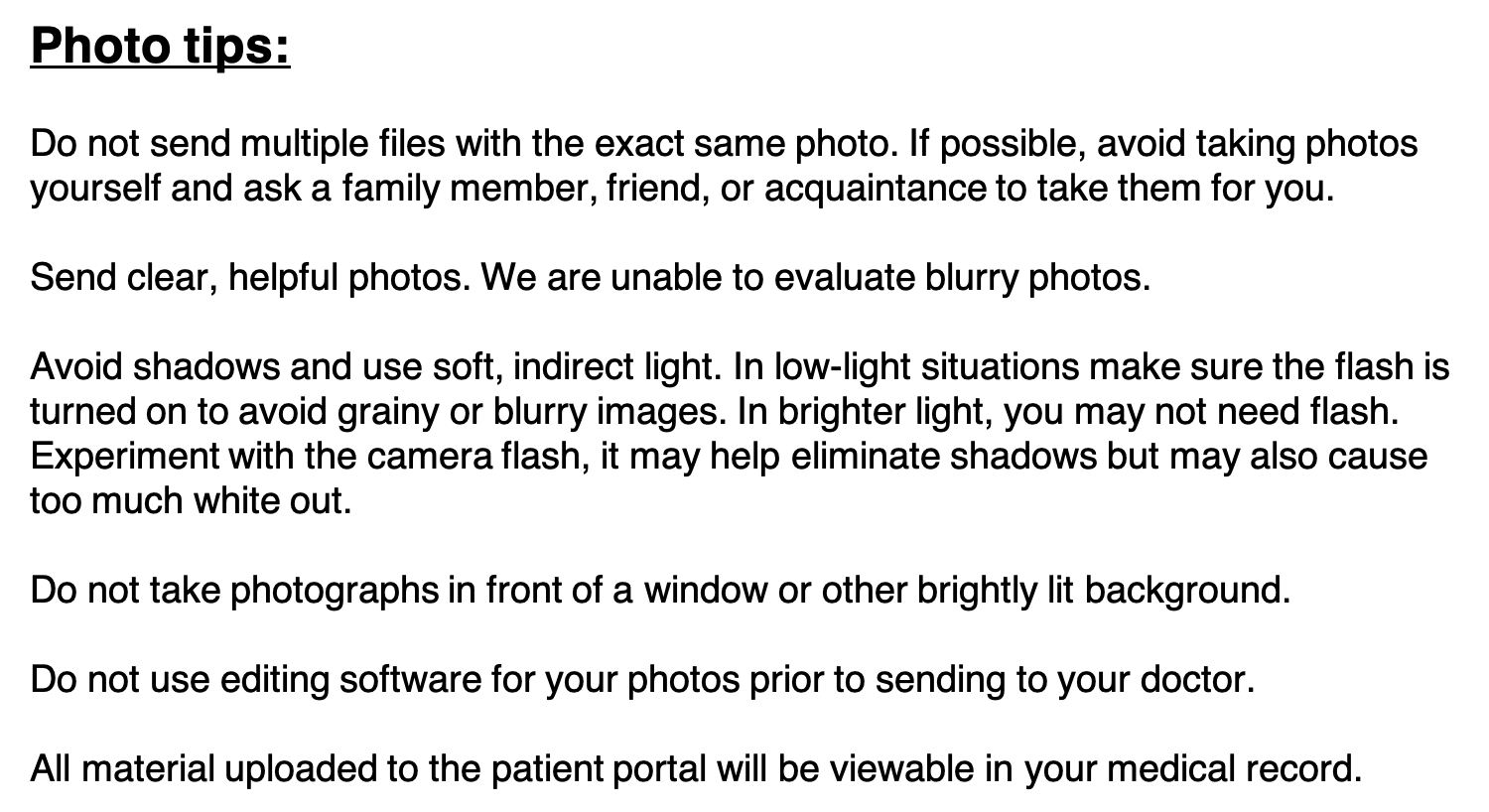}}
\caption{Example set of instructions given to patients at Stanford Health Care for how to take images for dermatology visits.}
\label{fig:telederm_instructions}
\end{figure}

Patients are guided on how to take photos of their lesions; see Figure~\ref{fig:telederm_instructions} for standard guidelines. Despite these instructions, it is common for patients to take blurry images, images in poor lighting conditions (e.g., too much glare or too dark), or images that do not adequately show the lesion (e.g., taken from too far away). Prior assessments of image quality in dermatology are not applicable to real world teledermatology, as trained medical professionals took the photos in these studies \cite{krupinski1999diagnostic}. So, we conducted an informal survey of dermatologists that suggests up to one fifth of all images sent in by patients could be of too low quality to be of use; see Table~\ref{tbl:derm_survey}.

Due to this high percentage of low quality images, dermatologists or other staff members screen images prior to a visit and request a patient to retake an image when necessary. This process is time consuming and can take a similar amount of time as a regularly scheduled visit. Moreover, it is common for patients to send in images just prior to a visit leaving no time for image quality screening. When these images are low quality, the clinical visit is spent coaching the patient on retaking the photo rather than the clinical issue. Therefore, poor quality images can significantly disrupt a clinician's schedule and affect clinical care.


We propose an automated machine learning method for assessing dermatology image quality and giving concrete feedback for how to improve quality when necessary (e.g., ``turn on camera flash" for dim lighting). We envision this solution as integrated into a smartphone application that can guide a patient through the process of taking an image with interactive, real-time feedback so that only high-quality photos are submitted for the televisit. This necessitates a computationally lightweight solution and motivates some of our design decisions.




 We detail our prototype algorithm, TrueImage, and assess our method on a dataset of dermatology photos as a proof-of-concept. The method provides a quality score to quantify how suitable a photo is for teledermatology. The score enables clinicians to flexibly set a threshold for filtering photo quality -- a more stringent threshold makes it more likely that the clinician works with high-quality images but could require more patients to retake photos. For example, we can reject $\sim$$50\%$ of poor quality images at while retaining $\sim$$80\%$ of good quality images; alternatively, we can reject $\sim$$10\%$ of all poor quality images while retaining $>$$95\%$ of good quality images.
 
 \paragraph{Contributions} We identify photo quality as an important emerging challenge for telehealth, especially for dermatology. There has been relatively little work in this area. We develop an algorithm for automatic quality detection in dermatology images and to provide guidance to patients. This can potentially improve the clinical workflow and efficiency.

\begin{table}[h]
\tbl{Results of a survey we conducted asking dermatologists how often patients send in poor quality photos. Samples size is 37. Several responders reported poor image frequency as high as $1/2$.}
{\begin{tabular}{@{}lcccr@{}}
\toprule
Frequency of poor quality photos per visit  & $1/5$ & $1/10$ & $1/20$ & $1/50$ \\ \colrule
Percent of survey response & 78.4\% & 10.8\% & 5.4\% & 5.4\% \\ \botrule
\end{tabular}}\label{tbl:derm_survey}
\end{table}

\section{Background}

Dermatology has become an important application of machine learning research in recent years with the success of deep learning and the acquisition of large dermatology datasets. Much of this work is related to disease diagnosis \cite{esteva2017dermatologist, han2020augmented} or lesion segmentation \cite{goyal2017multi}, and most public data is taken using dermatoscopes, a special tool for magnifying lesions. However, as large-scale teledermatology is relatively new, little work has been done in solving problems specific to automatically assessing the quality of patient-taken images. There are several related problems that we detail here. 

\subsection{Clinical Image Guidelines}

\begin{figure}[h]
\centerline{\includegraphics[width=14cm]{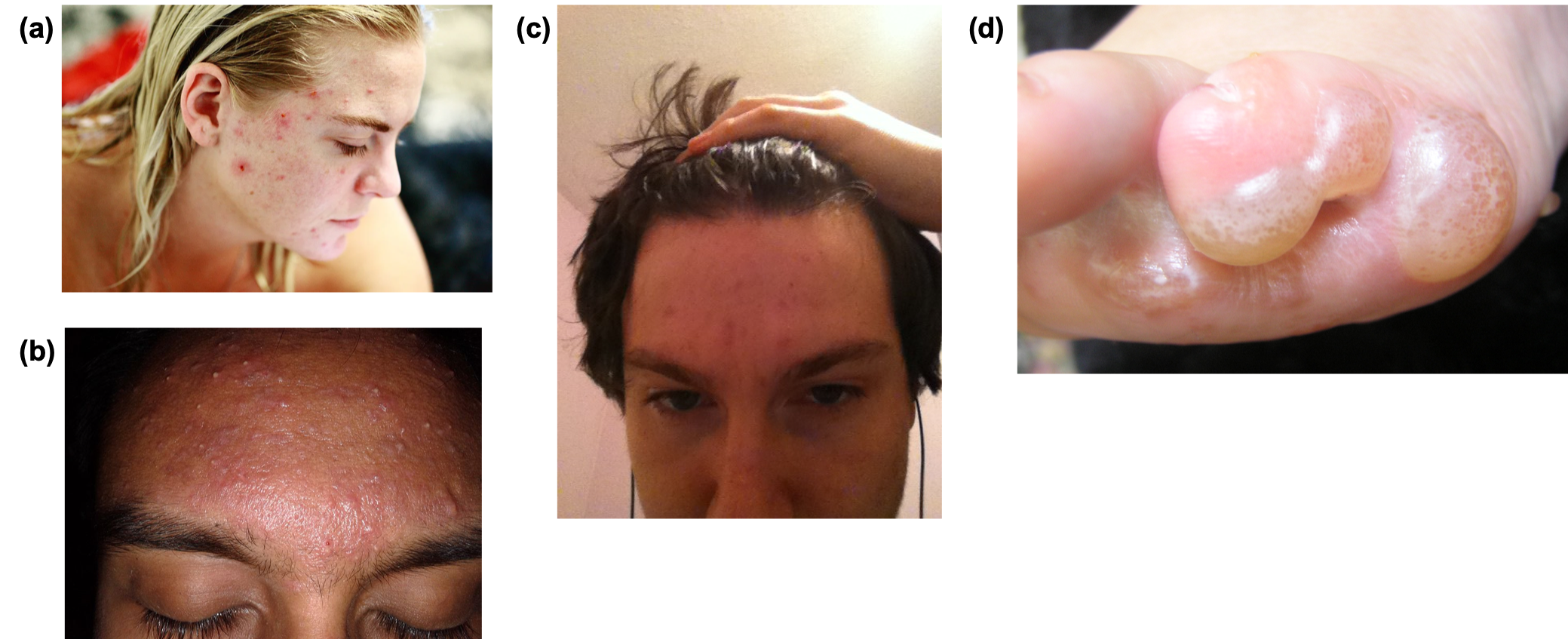}}
\caption{Examples of poor quality images. (a) is a good quality image; the acne is in-focus, the lighting is not too dark and there is little glare. (b) has excessive glare and shadows on the skin. (c) is too blurry. Note that the lighting is also dim here; brighter lighting would likely also reduce issues with blur. (d) is cropped and zoomed in too close; the slight blur and glare exacerbates the problem.}
\label{fig:image_examples}
\end{figure}

Photography for dermatology is commonly used in a clinical setting for both educational purposes and to track disease progression in patients. To ensure high quality photos, there are several guidelines that have been developed to counter the common issues that produce low quality photos in dermatology \cite{muraco2020improved}. See Figure~\ref{fig:image_examples} for illustrative examples. These issues can be summarized as
\begin{enumerate}
  \item[1.] Skin lesion area is blurry (Figure~\ref{fig:image_examples}c).
  \item[2.] Skin is discolored due to lighting conditions -- this may be induced by a dim environment, excessive shadows, excessive glare (e.g., due to camera flash), or the background reflecting tinted light (Figure~\ref{fig:image_examples}b).
  \item[3.] Skin lesion is cropped or image is taken from too far away (Figure~\ref{fig:image_examples}d).
  \item[4.] Image is distorted due to camera effects (e.g., fish-eye effect).
  \item[5.] Background is distracting or patient is wearing distracting clothing and jewelry.
  \item[6.] Image is taken at a poor orientation (e.g., a leg is photographed horizontally; a vertical photograph is preferable so that the entire frame is filled with the leg).
\end{enumerate}

Items 5 and 6 above tend to be less critical for dermatologists to make an assessment. Empirically, we have noted that the most common issues in patient-taken images are items 1-3, with the ordering corresponding to their prevalence. So in this work, we focus on these 3 issues.

\subsection{Image Quality Assessment}

Image quality assessment (IQA) methods attempt to measure the quality of digital images, where image quality is usually defined by human labelers. IQA methods can be split into two categories -- full-reference IQA (FR-IQA) which require a high quality reference image and no-reference IQA (NR-IQA) which work given a single image; some techniques are adaptable to both settings \cite{bosse2017deep}. Additionally, some methods are designed with respect to specific distortions like Gaussian blur, additive white noise, and JPEG compression artifacts \cite{liu2008image, li2015no, liu2017rankiqa}, while others are general purpose and adapt to the distortions present in a training dataset \cite{zhang2015feature}. IQA has generally been studied for use on natural images, though some techniques have been adapted for detecting artifacts in MRI, CT, and ultrasonography \cite{chow2016review} in clinical settings.

Due to the cost of labeling data (trained human labelers are required), most IQA datasets are small. However, deep learning methods have become prevalent recently and typically utilize data augmentation techniques like applying fixed distortions (e.g., Gaussian blur) to high quality images \cite{liu2017rankiqa}, utilizing generative models like GANs \cite{lin2018hallucinated}, or through leveraging larger image classification datasets for transfer learning \cite{gao2018blind}. 

Classical methods also have good performance, in particular when a specific, known distortion is in consideration. Of particular interest to us, blur detection is a well-studied problem and efficient classical algorithms exist \cite{liu2008image, pertuz2013analysis}. These algorithms generally rely on detecting the magnitude of low and high-frequency content in images -- blurry images tend to have reduced high-frequency content.

\subsection{Semantic Segmentation}

Semantic segmentation is the problem of generating per-pixel class labels in an image. In our case, we are interested in a binary semantic segmentation problem of labeling skin and non-skin pixels in an image; segmentation is important to us as we are interested in the quality of \textit{only} the parts of the image containing the lesion (e.g., the background can be blurry, but the lesion cannot). 

Deep learning methods have dominated the field in recent years, with most modern methods relying on fully convolutional networks \cite{guo2018review, ulku2019survey}. Methods like Mask RCNN \cite{he2017mask} and YOLACT \cite{bolya2019yolact} layer a semantic segmentation module over an object detection framework to allow for instance level semantic segmentation. Other techniques use recurrent connections when multiple images (e.g., a video sequence) are available \cite{nilsson2018semantic}.

Classical methods also exist and utilize a variety of approaches. Of particular interest to us, there are a large number of methods designed specifically for skin detection as part of gesture recognition or facial detection algorithms. Some of the most simple methods utilize decision trees learned from a dataset of skin pixels and classify each pixel independently \cite{bhatt2009efficient, kolkur2017human}; other methods fit the distribution or skin colors using, for example, histogram-based techniques or a Gaussian mixture model and apply a threshold on the predicted probability to classify skin pixels \cite{jones2002statistical}. More sophisticated methods apply additional steps to account for spatial information in the image \cite{mahmoodi2017fast}.

\section{TrueImage Algorithm}

\begin{figure}[h]
\centerline{\includegraphics[width=14cm]{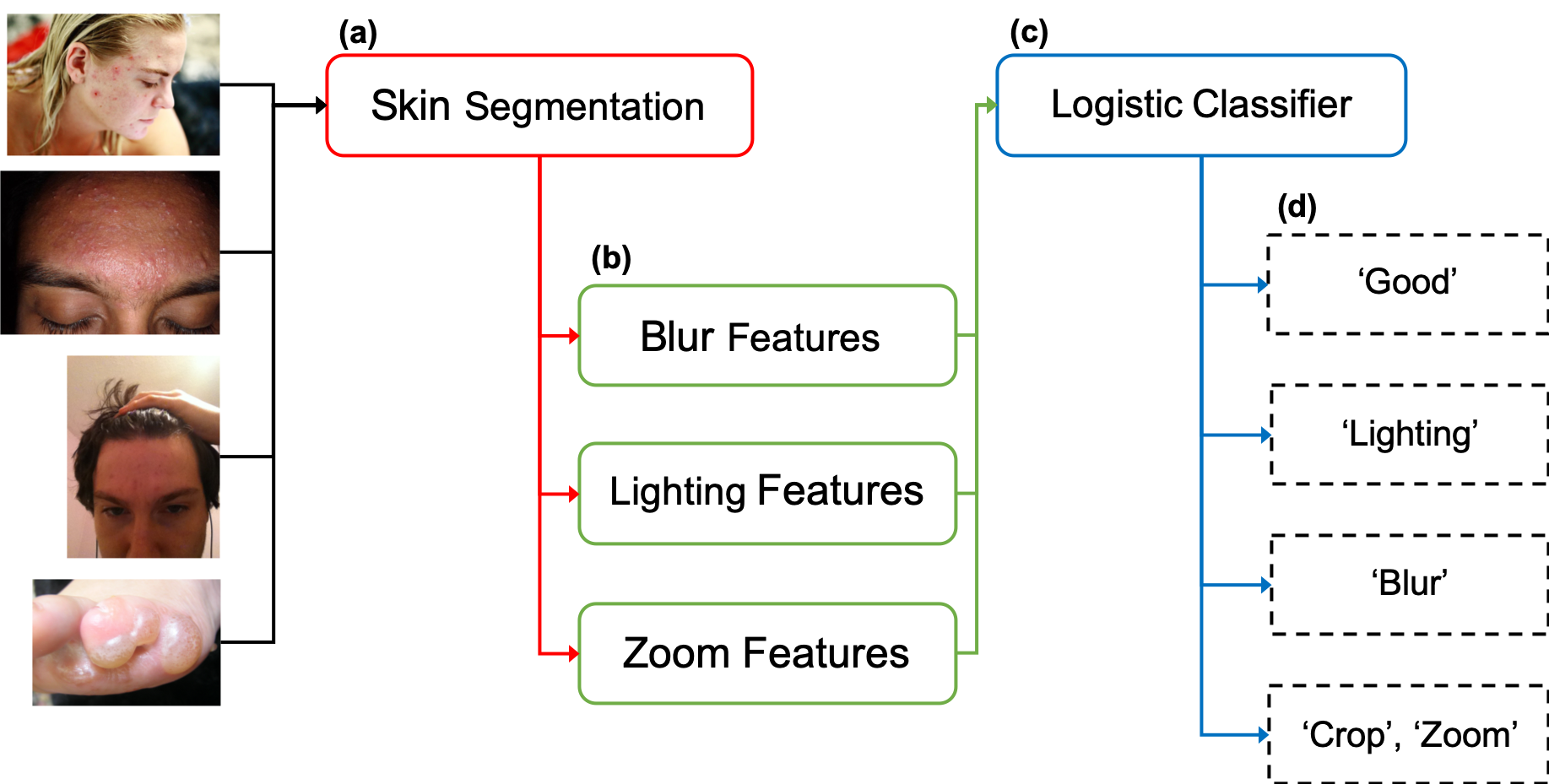}}
\caption{Workflow schematic of the TrueImage dermatology image quality detection algorithm. (a) Image is input for skin segmentation. (b) After segmentation, the original image and segmentation mask are used for feature generation in three groups. Principal component analysis (PCA) is used to reduce the dimensionality of features within each feature group. (c) 4 classifiers are applied to the concatenated feature vectors, giving us (d) labels for reason(s) of poor image quality.}
\label{fig:algorithm_pipeline}
\end{figure}

Our algorithm can be described in 3 stages as shown in Figure~\ref{fig:algorithm_pipeline}. It consists of (1) semantic segmentation to identify skin regions, (2) feature generation, and (3) a quality classifier applied to these features. Our algorithm design is guided by two considerations/constraints:

\begin{enumerate}
    \item[1.] We desire a computationally lightweight algorithm, due to the end-goal of having an interactive system deployable on older-generation smartphones.
    \item[2.] As labeling data is costly, a more data-efficient algorithm is preferable.
  \end{enumerate}

These constraints motivated us to represent each photo by a relatively small number of interpretable features, which we explain in detail below.
Additionally, the algorithm itself is more interpretable and so we can more easily ensure it is robust to various skin tones.

\subsection{Semantic Segmentation}

We use a per-pixel semantic segmentation algorithm to identify skin and lesion pixels. Each pixel is classified independently. Each pixel is transformed from RGB into both YCrCb and HSV color spaces, and the two representations are concatenated giving a vector in $\mathbb{R}^6$. We then use a Gaussian mixture model to assign the pixel a score corresponding to its likelihood of being ``skin"; applying a threshold to the pixels scores gives us our semantic segmentation.

Additionally, we always consider border pixels to be non-skin. Empirically, we have seen that patient-taken images are generally well-centered. Thus, center cropping is a simple way of ensuring we are assessing the lesion area and not the surrounding skin or background clutter. 

We also implement a simple, per-pixel lesion segmentation algorithm. The algorithm takes a pixel, transforms it to the LAB color space, and keeps the brightest pixels from each color channel. Then we compute the fraction of these brightest pixels located near the center of the image. The color channel with the highest fraction is used as the lesion mask; we additionally perform a bitwise-and operation with the skin segmentation mask to reduce false positives. An inspection of 15 images suggested that this algorithm detects red and dark patches well.

Although threshold-based algorithms are generally inferior to modern deep learning algorithms, our algorithm performs adequately in our setting, largely due to the constraints on our distribution of images (e.g., images are generally centered). Furthermore, since our end goal is a downstream task that uses the segmentation as an input, slight errors do not have significant impact.

\subsection{Features}

We consider 3 major reasons for poor quality images: blur, lighting conditions, and zoom. Although there are other reasons as noted in \cite{muraco2020improved}, we have found empirically that these 3 are the most common issues in patient-taken images. We generate features designed to capture good-to-bad image variance for each of these issues. 

\subsubsection{Blur Features}

Blur can be characterized by a higher presence of low-frequency components in an image as well as other effects like decreased color saturation. These effects can be measured by computation of the Fourier Transform on an image, through analysis of the image gradient (or higher order derivatives), or through various other means like looking at the color saturation \cite{liu2008image, pertuz2013analysis}. 

Subsequent to our semantic segmentation, we uniformly sample $100$ $32 \times 32$ pixel patches that contain at least $90\%$ ``skin" pixels. For each patch $P$, we compute 

\begin{enumerate}
  \item[1.] the magnitude of the high-pass filtered patch, 
      $$ \frac{1}{32^2} \sum_{i,j \in [32]} 20 \ln \left( || F_h(P)_{i,j} || \right), $$ 
    where $F_h$ denotes the high-pass filter, and
  \item[2.] the variance of the Laplacian of the patch, 
      $$ \frac{1}{32^2} \sum_{i,j \in [32]} (L(x,y) - \mathbb{E}[L(x,y)])^2, $$ 
    where $L(x,y) = \frac{\partial^2 P}{\partial x^2} + \frac{\partial^2 P}{\partial y^2}$.
\end{enumerate}

Subsequently, we summarize the patch-level distribution by computing the mean, median, max, min, and standard deviation of each feature across patches. This process results in $10$ total features. We apply principal component analysis (PCA) to these features and keep the top $5$ principal components.

\subsubsection{Lighting Features}

To detect poor lighting conditions, we use two types of features. We compute these features per patch, using the same $100$ patches as for blur.

\begin{enumerate}
  \item[1.] We transform the image to grayscale, $G$, and compute per each patch $P_G$ (1) $ underexposed := P_G[P_G < 50] $ and (2) $ overexposed := P_G[P_G > 205] $. Color values are integers in the range $[0, 255]$. $underexposed$ assess the amount of shadows and dim lighting in the image while $overexposed$ assess the amount of glare. Note that both $underexposed$ and $overexposed$ are sets of values; we summarize each set by computing the median and the upper and lower quartiles. We summarize the patch-level distribution by computing the mean, median, max, min, and standard deviation across patches for each feature, resulting in 30 features.
  \item[2.] We consider the probability distribution given by the Gaussian mixture model trained for skin segmentation for each patch. This distribution gives us information on the glare and shadow content, as well as discoloration in the skin due to poor lighting (e.g., a blue tinted light). We compute the median and lower and upper quartiles per patch, and subsequently compute the mean, median, max, min, and standard deviation across patches.
\end{enumerate}

The above process results in 45 total features; we reduce this to $5$ features using PCA.

\subsubsection{Zoom Features}

For assessing crop and zoom, we compute the ratio of skin to non-skin pixels and lesion to non-lesion detected inside the center cropped area. Images zoomed out too far will have fewer ``skin" pixels near their center. Similarly, a high fraction of lesion pixels near the boundary of an image suggest that the lesion is not shown with adequate context.

Note that the relevance of these features are entirely dependent on the skin and lesion segmentation algorithms; to counteract deficiencies in our segmentation algorithm, we apply a more generous threshold for computing these features. 

\subsection{Feature Aggregation}

We concatenate across feature groups and train four binary classifiers using logistic regression: one predicts whether an image is good quality, and the remaining three predict whether the image is blurry, has poor lighting, or is zoomed out / cropped respectively.

\section{Training and Evaluation}

\subsection{Skin Segmentation Model}

The Gaussian mixture model is trained to fit the distribution of skin pixels using the dataset from Bhatt et. al \cite{bhatt2009efficient}. This dataset consists of roughly 50000 skin pixels sampled from 14,701 face images of a diverse set of individuals across age, gender, and ethnicity. Our model is fit using the standard expectation-maximization algorithm. Note that though this model is trained with pixels from face images, it is able to identify skin pixels in a generalizable manner. 


\subsection{Logistic Classifiers}

To train our binary classifiers, we use four datasets. The first dataset was curated from Google Images using images licensed for free commercial use and contains images of various diseases and lesion types. A dermatologist added labels assessing image quality and giving reasons for poor quality. Note that the assigned labels should be interpreted as ``too blurry to make an assessment" (if the assigned label is blurry). We augmented this dataset by applying one of two types of distortions to all good quality images. We split the dataset into training, validation, and test sets prior to augmentation so as to avoid data leakage.
\begin{enumerate}
    \item[1.] (blur) Gaussian blur with randomly sized kernel.
    \item[2.] (zoom/crop) Select random corner cropping of the image.
\end{enumerate}
A dermatologist separately labeled some of the artificial images for use in our test set; only images the dermatologist deemed realistic were included for testing. See rows 1 and 2 (``Web" and ``Web-Augmented") of Table~\ref{tbl:derm_dataset} for a breakdown of the labels in this data.

We also used a dataset collected at Stanford Health Care. This dataset contains images taken by a dermatologist of various lesions observed in clinic. See row 3 (``Stanford") of Table~\ref{tbl:derm_dataset} for details. We do not augment this dataset.

As these datasets have relatively few images of poor lighting conditions, we additionally collected our own poor lighting images by taking pictures of the authors and their families with their consent. We used this data only for training purposes. See row 4 (``Extra") of Table~\ref{tbl:derm_dataset} for details. By aggregating across these four datasets from heterogeneous sources, we provide more representative photos to train and evaluate our approach.

\begin{table}[h]
\tbl{Image counts for our dermatology dataset. Some images have multiple labels. Images are randomly split into train, validation and test sets. 
}
{\begin{tabular}{@{}lccccr@{}}
\toprule
Data source & Total & Good & Blurry & Poor Lighting & Poor Zoom / Crop \\ \colrule
Web & 55 & 46 & 5 & 5 & 1 \\ 
Web-Augmented & 179 & 14 & 80 & 0 & 85 \\
Stanford & 99 & 86 & 5 & 7 & 2 \\
Extra & 29 & 0 & 0 & 29 & 0 \\ \colrule
All data & 362 & 146 & 90 & 41 & 88 \\ \botrule
\end{tabular}}\label{tbl:derm_dataset}
\end{table}

\subsection{Results on Dermatology Dataset}

\begin{figure}[h]
\centerline{\includegraphics[width=14cm]{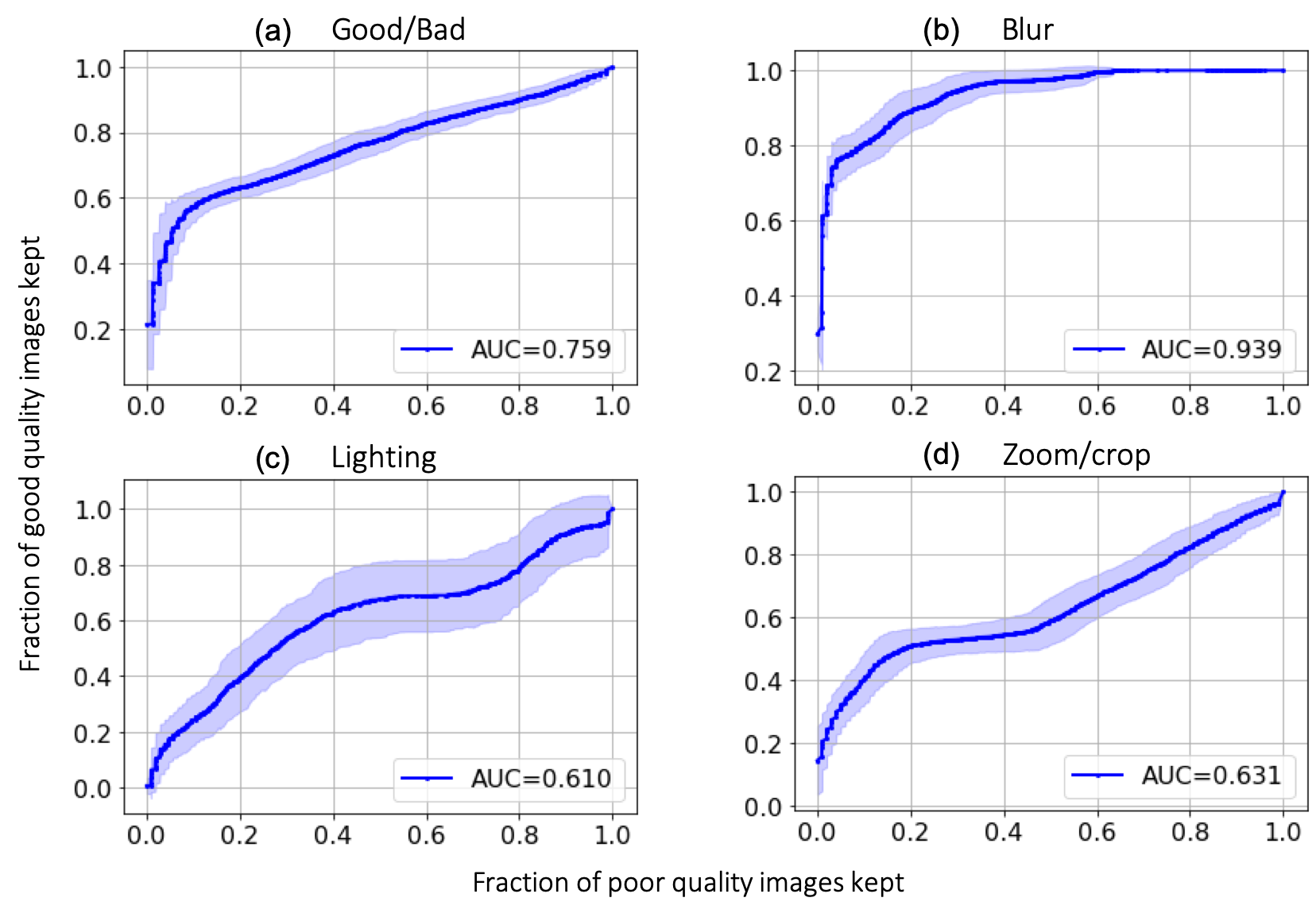}}
\caption{Receiver operating characteristic (ROC) curve of TrueImage; 1 standard deviation confidence interval shaded in. Classifiers are for (a) Good/bad, (b) Blur, (c) Lighting, (d) Zoom/crop.}
\label{fig:results}
\end{figure}

After training our algorithm, we evaluated it on the test dataset described above. Our results are shown in Figure~\ref{fig:results}. There are four plots shown, one for each of the labels we generate: (a) good/poor quality, (b) blurry/not blurry, (c) good/poor lighting, (d) good/poor zoom or crop. We plot the receiver operating characteristic (ROC) curve.

Looking at Figure~\ref{fig:results}a, we notice that (1) we can reject $\sim$$50\%$ of poor quality images while retaining $\sim$$80\%$ of good quality images or (2) we can reject $\sim$$10\%$ of poor quality images while retaining $>$$95\%$ of good quality images; this is particularly important as it suggests we can set TrueImage's parameters to reduce time spent by clinicians without adversely affecting patients. Looking at Figure~\ref{fig:results}b-d, we see that we can detect blur quite well, but can only detect poor lighting or crop moderately well. Lastly, we note that the result in Figure~\ref{fig:results}a is dependent on the distribution of poor quality images. Empirically, we have found that blurry images are the most common followed by poor lighting conditions.

Many of the ``good" quality images are not actually good quality, but are adequate enough for clinical assessment. This label may vary between dermatologists, and the threshold parameter should be tuned per dermatologist and even per disease category (e.g., inflammatory lesions vs. non-inflammatory lesions). We envision that individual dermatologists or group practices will set their thresholds based on their preferences prior to sharing the application with patients; moreover, these settings would be adjustable by clinicians based on patient feedback. Some images are in a gray-area where their image quality rating is situation dependent (e.g., what context the dermatologist is viewing them in). The $\sim$$10\%$ we can reject outright are in a set of images that are indisputably poor quality. 

\section{Discussion and Future Work}

This work highlights photo quality as a significant and understudied challenge for teledermatology. We develop a novel, automated approach to detect poor quality dermatology photos. 

\begin{figure}[h]
\centerline{\includegraphics[width=14cm]{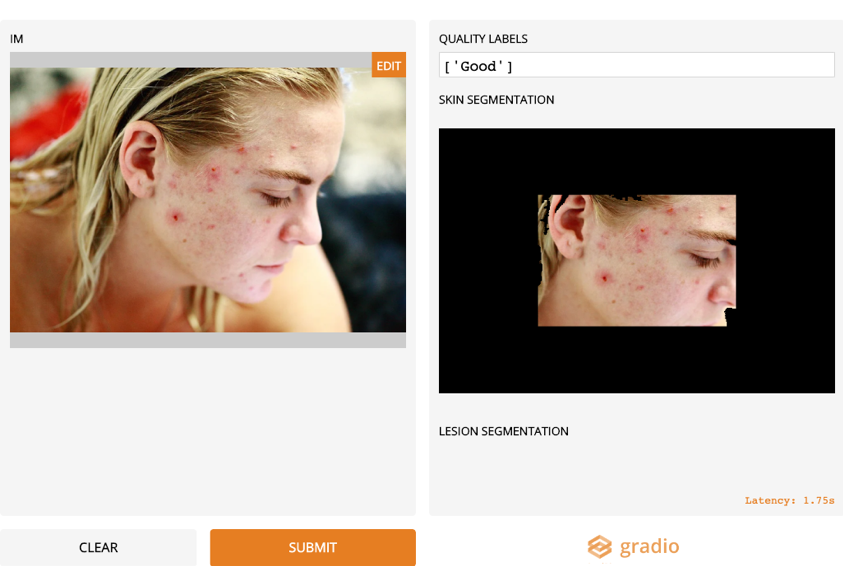}}
\caption{User interface design using gradio \cite{abid2020online}. Displays input image, skin segmentation, and quality classification. 
}
\label{fig:current_interface}
\end{figure}

We have also implemented an interactive user interface for TrueImage using gradio \cite{abid2020online} shown in Figure~\ref{fig:current_interface}, which will facilitate usage of TrueImage in clinical pilot studies. The eventual goal is to make an interactive interface run by the user to guide them in real-time in taking clinical photos. Our algorithm is computationally very efficient, and an unoptimized, single-threaded implementation takes about $1$ second to run per photo using a standard laptop with a 1.8 GHz Intel Core i5 processor. This timing was computed by averaging over the runtime of $20$ images.
Because teledermatology is still quite new, there are limited publicly available datasets of patient photos annotated with quality measures. To address this challenge, we curated a dataset of 362 images from diverse sources and annotated the quality of each photo along with reasons for poor quality images.
While our dataset is a good first step, creating larger datasets of dermatologist labeled photos would be a useful resource for telehealth research and can further improve the performance of TrueImage.  Moreover, with a larger dataset, we hope to create more specific feedback for patients (e.g. instead of saying "poor lighting", stating "lighting is too dark" or "lighting is too bright"). 
Additionally, most of the images are of patients with lighter skin tones. A larger dataset with more diverse skin types is critical for TrueImage to be more broadly useful. However, our current results demonstrate that it is possible to robustly detect image quality in dermatology photos. 

\bibliographystyle{ws-procs11x85}
\bibliography{main}

\begin{thebibliography}{10}

\bibitem{mehrotra2020impact}
A.~Mehrotra, M.~Chernew, D.~Linetsky, H.~Hatch and D.~Cutler, The impact of the
  covid-19 pandemic on outpatient visits: Practices are adapting to the new
  normal  (2020).

\bibitem{krupinski1999diagnostic}
E.~A. Krupinski, B.~LeSueur, L.~Ellsworth, N.~Levine, R.~Hansen, N.~Silvis,
  P.~Sarantopoulos, P.~Hite, J.~Wurzel, R.~S. Weinstein {\em et~al.},
  Diagnostic accuracy and image quality using a digital camera for
  teledermatology, {\em Telemedicine Journal} {\bf 5}, 257  (1999).

\bibitem{esteva2017dermatologist}
A.~Esteva, B.~Kuprel, R.~A. Novoa, J.~Ko, S.~M. Swetter, H.~M. Blau and
  S.~Thrun, Dermatologist-level classification of skin cancer with deep neural
  networks, {\em nature} {\bf 542}, 115  (2017).

\bibitem{han2020augmented}
S.~S. Han, I.~Park, S.~E. Chang, W.~Lim, M.~S. Kim, G.~H. Park, J.~B. Chae,
  C.~H. Huh and J.-I. Na, Augmented intelligence dermatology: Deep neural
  networks empower medical professionals in diagnosing skin cancer and
  predicting treatment options for 134 skin disorders, {\em Journal of
  Investigative Dermatology}   (2020).

\bibitem{goyal2017multi}
M.~Goyal and M.~H. Yap, Multi-class semantic segmentation of skin lesions via
  fully convolutional networks, {\em arXiv preprint arXiv:1711.10449}   (2017).

\bibitem{muraco2020improved}
L.~Muraco, Improved medical photography: key tips for creating images of
  lasting value, {\em JAMA dermatology} {\bf 156}, 121  (2020).

\bibitem{bosse2017deep}
S.~Bosse, D.~Maniry, K.-R. M{\"u}ller, T.~Wiegand and W.~Samek, Deep neural
  networks for no-reference and full-reference image quality assessment, {\em
  IEEE Transactions on Image Processing} {\bf 27}, 206  (2017).

\bibitem{liu2008image}
R.~Liu, Z.~Li and J.~Jia, Image partial blur detection and classification, in
  {\em 2008 IEEE conference on computer vision and pattern recognition\/},
  (IEEE, 2008).

\bibitem{li2015no}
L.~Li, W.~Lin, X.~Wang, G.~Yang, K.~Bahrami and A.~C. Kot, No-reference image
  blur assessment based on discrete orthogonal moments, {\em IEEE transactions
  on cybernetics} {\bf 46}, 39  (2015).

\bibitem{liu2017rankiqa}
X.~Liu, J.~van~de Weijer and A.~D. Bagdanov, Rankiqa: Learning from rankings
  for no-reference image quality assessment, in {\em Proceedings of the IEEE
  International Conference on Computer Vision\/},  (IEEE, 2017).

\bibitem{zhang2015feature}
L.~Zhang, L.~Zhang and A.~C. Bovik, A feature-enriched completely blind image
  quality evaluator, {\em IEEE Transactions on Image Processing} {\bf 24}, 2579
   (2015).

\bibitem{chow2016review}
L.~S. Chow and R.~Paramesran, Review of medical image quality assessment, {\em
  Biomedical signal processing and control} {\bf 27}, 145  (2016).

\bibitem{lin2018hallucinated}
K.-Y. Lin and G.~Wang, Hallucinated-iqa: No-reference image quality assessment
  via adversarial learning, in {\em Proceedings of the IEEE Conference on
  Computer Vision and Pattern Recognition\/},  (IEEE, 2018).

\bibitem{gao2018blind}
F.~Gao, J.~Yu, S.~Zhu, Q.~Huang and Q.~Tian, Blind image quality prediction by
  exploiting multi-level deep representations, {\em Pattern Recognition} {\bf
  81}, 432  (2018).

\bibitem{pertuz2013analysis}
S.~Pertuz, D.~Puig and M.~A. Garcia, Analysis of focus measure operators for
  shape-from-focus, {\em Pattern Recognition} {\bf 46}, 1415  (2013).

\bibitem{guo2018review}
Y.~Guo, Y.~Liu, T.~Georgiou and M.~S. Lew, A review of semantic segmentation
  using deep neural networks, {\em International journal of multimedia
  information retrieval} {\bf 7}, 87  (2018).

\bibitem{ulku2019survey}
I.~Ulku and E.~Akagunduz, A survey on deep learning-based architectures for
  semantic segmentation on 2d images, {\em arXiv preprint arXiv:1912.10230}
  (2019).

\bibitem{he2017mask}
K.~He, G.~Gkioxari, P.~Doll{\'a}r and R.~Girshick, Mask r-cnn, in {\em
  Proceedings of the IEEE international conference on computer vision\/},
  (IEEE, 2017).

\bibitem{bolya2019yolact}
D.~Bolya, C.~Zhou, F.~Xiao and Y.~J. Lee, Yolact: Real-time instance
  segmentation, in {\em Proceedings of the IEEE international conference on
  computer vision\/},  (IEEE, 2019).

\bibitem{nilsson2018semantic}
D.~Nilsson and C.~Sminchisescu, Semantic video segmentation by gated recurrent
  flow propagation, in {\em Proceedings of the IEEE Conference on Computer
  Vision and Pattern Recognition\/},  (IEEE, 2018).

\bibitem{bhatt2009efficient}
R.~B. Bhatt, G.~Sharma, A.~Dhall and S.~Chaudhury, Efficient skin region
  segmentation using low complexity fuzzy decision tree model, in {\em 2009
  Annual IEEE India Conference\/},  (IEEE, 2009).

\bibitem{kolkur2017human}
S.~Kolkur, D.~Kalbande, P.~Shimpi, C.~Bapat and J.~Jatakia, Human skin
  detection using rgb, hsv and ycbcr color models, {\em arXiv preprint
  arXiv:1708.02694}   (2017).

\bibitem{jones2002statistical}
M.~J. Jones and J.~M. Rehg, Statistical color models with application to skin
  detection, {\em International Journal of Computer Vision} {\bf 46}, 81
  (2002).

\bibitem{mahmoodi2017fast}
M.~R. Mahmoodi, Fast and efficient skin detection for facial detection, {\em
  arXiv preprint arXiv:1701.05595}   (2017).

\bibitem{abid2020online}
A.~Abid, A.~Abdalla, A.~Abid, D.~Khan, A.~Alfozan and J.~Zou, An online
  platform for interactive feedback in biomedical machine learning, {\em Nature
  Machine Intelligence} {\bf 2}, 86  (2020).

\end{thebibliography}

\paragraph{Data used in this study}
Public data used in this study was released under a Wikimedia Commons-acceptable license (i.e., free to use for any purpose). The Stanford data comes from the Stanford Dermatology Clinic and is covered by IRB protocol 57673.

\end{document}